\documentclass[acmtog,nonacm]{acmart}

\usepackage{fp}
\usepackage{expl3}[2012-07-08]
\ExplSyntaxOn
\ExplSyntaxOff

\usepackage{amsmath,amsfonts,bm}
\usepackage{bm}

\def\x{{x}}

\def\xi{{\x_i}}

\newcommand{\ignorethis}[1]{}

\def\eqref#1{equation~\ref{#1}}

\def\1{\bm{1}}

\DeclareMathAlphabet{\mathsfit}{\encodingdefault}{\sfdefault}{m}{sl}
\SetMathAlphabet{\mathsfit}{bold}{\encodingdefault}{\sfdefault}{bx}{n}

\newcommand{\ignore}[1]{}

\makeatletter
\DeclareRobustCommand\onedot{\futurelet\@let@token\@onedot}
\def\@onedot{\ifx\@let@token.\else.\null\fi\xspace}

\makeatother

\usepackage{multirow}
\usepackage{booktabs} 
\usepackage{url}
\usepackage{stfloats}
\usepackage{amsmath,bm}
\usepackage[ruled]{algorithm2e} 

\SetAlFnt{\small}
\SetAlCapFnt{\small}
\SetAlCapNameFnt{\small}
\SetAlCapHSkip{0pt}
\setcopyright{none}
\usepackage{overpic} 
\usepackage{subcaption}

\DeclareMathAlphabet{\mathcal}{OMS}{cmsy}{m}{n}

\usepackage{array}
\newcolumntype{L}[1]{>{\raggedright\let\newline\\\arraybackslash\hspace{0pt}}m{#1}}
\newcolumntype{C}[1]{>{\centering\let\newline\\\arraybackslash\hspace{0pt}}m{#1}}
\newcolumntype{R}[1]{>{\raggedleft\let\newline\\\arraybackslash\hspace{0pt}}m{#1}}

\newcommand{\sysName}{IconShop}

\begin{document}
\title{IconShop: Text-Guided Vector Icon Synthesis with Autoregressive Transformers}

\author{Ronghuan Wu}
\affiliation{
  \institution{City University of Hong Kong}
}
\email{ronghwu2-c@my.cityu.edu.hk}
\author{Wanchao Su}
\affiliation{
  \institution{City University of Hong Kong}
}
\email{wanchao_su@outlook.com}
\author{Kede Ma}
\affiliation{
  \institution{City University of Hong Kong}
}
\email{kede.ma@cityu.edu.hk}
\author{Jing Liao$^*$}\thanks{*Corresponding Author}
\affiliation{
  \institution{City University of Hong Kong}
}
\email{jingliao@cityu.edu.hk}

\begin{abstract}
Scalable Vector Graphics (SVG) is a popular vector image format that offers good support for interactivity and animation. Despite its appealing characteristics, creating custom SVG content can be challenging for users due to the steep learning curve required to understand SVG grammars or get familiar with professional editing software. Recent advancements in text-to-image generation have inspired researchers to explore vector graphics synthesis using either image-based methods (i.e., text $\rightarrow$ raster image $\rightarrow$ vector graphics) combining text-to-image generation models with image vectorization, or language-based methods (i.e., text $\rightarrow$ vector graphics script) through pretrained large language models. Nevertheless, these methods suffer from limitations in terms of generation quality, diversity, and flexibility.
In this paper, we introduce \sysName, a text-guided vector icon synthesis method using autoregressive transformers. The key to success of our approach is to sequentialize and tokenize SVG paths (and textual descriptions as guidance) into a uniquely decodable token sequence. With that, we are able to exploit the sequence learning power of autoregressive transformers, while enabling both unconditional and text-conditioned icon synthesis. Through standard training to predict the next token on a large-scale vector icon dataset accompanied by textural descriptions, the proposed \sysName~consistently exhibits better icon synthesis capability than existing image-based and language-based methods both quantitatively (using the FID and CLIP scores) and qualitatively (through formal subjective user studies). Meanwhile, we observe a dramatic improvement in generation diversity, which is validated by the objective Uniqueness and Novelty measures. More importantly, we demonstrate the flexibility of \sysName~with multiple novel icon synthesis tasks, including icon editing, icon interpolation, icon semantic combination, and icon design auto-suggestion. The project page is {\color{blue}\url{https://icon-shop.github.io/}}.
\end{abstract}

%
%

\begin{teaserfigure}
  \includegraphics[width=\textwidth]{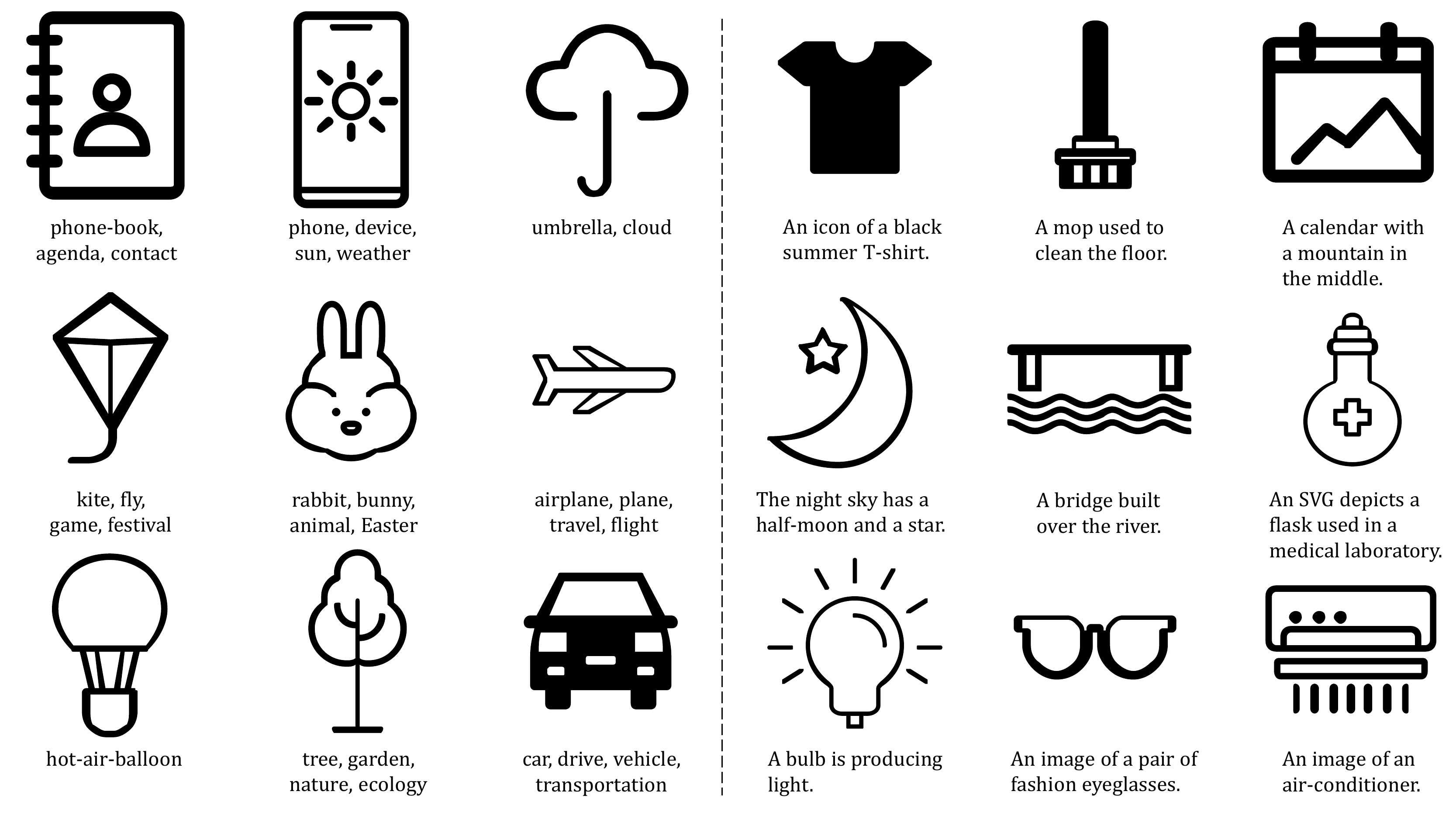}
  \caption{Vector icons generated with text prompts. The proposed \sysName~supports vector icon synthesis from keywords (left panel) and natural phrases and sentences (right panel).}
  \label{fig:teaser}
\end{teaserfigure}


\maketitle

\section{Introduction}
As a form of computer graphics, vector graphics represents visual content based directly on geometric shapes (via command lines and arguments), and is widely used in scientific and artistic applications, including architecture, surveying, 3D rendering, typography, and graphic design. Compared with raster graphics, vector graphics is preferred when a high degree of geometric precision is required across arbitrary scales, among which Scalable Vector Graphics (SVG) is a popular vector graphics file format, particularly in creative industries. Generally, creating SVG content is difficult for non-professional users. It is tedious and time-consuming to gain adequate knowledge of SVG grammars and/or master professional editing software such as Adobe Illustrator. Recently, there have been impressive successes in generating raster images from text, providing a convenient and efficient means of fulfilling users' image design intents. Thus, it is highly desirable to build a computational system that can accomplish something similar in the field of SVG, allowing accurate and flexible SVG content synthesis guided by intuitive textual descriptions.

One straightforward approach is adapting raster image generation to vector synthesis by converting the image outputs of text-to-image generation models into vector graphics using image vectorization methods. While such image-based methods (i.e., text $\rightarrow$ raster image $\rightarrow$ vector graphics) directly incorporate recent advances in text-to-image generation (e.g., Stable Diffusion~\cite{rombach2022high}) into SVG generation, their results are often unsatisfactory. This is because text-to-image models are mostly trained to generate complex natural geometric shapes and color appearances, and are less probable to reproduce SVG styles with simple geometric primitives and flat colors. Furthermore, to fit such complex raster images, vectorization methods often use jagged paths with unwanted corners and crossovers, resulting in visible and annoying artifacts.

As SVG is based on Extensible Markup Language (XML), one research avenue for synthesizing SVG content from text is to train Sequence-To-Sequence (seq2seq) models that take text prompts as input and directly produce SVG scripts as output. We refer to these as language-based methods (i.e., text $\rightarrow$ vector graphics script). 
Despite the conceptual simplicity, SVG involves complex grammars, and naive tokenization of SVG as a form of natural language may result in complex and lengthy token sequences, which complicate subsequent probabilistic modeling. Preliminary experiments~\cite{bubeck2023sparks} show that Large Language Models (LLMs) like GPT-4~\cite{openai2023gpt4} tend to combine basic geometric shapes, such as \verb|Circle| and \verb|Ellipse|, to convey semantic information with relatively good text-SVG alignment. However, the synthesis results show limited complexity and diversity (see Section \ref{experiment_comparison}), which is inadequate for real-world SVG applications.

In this paper, we develop an autoregressive transformer-based method that supports accurate and flexible SVG content synthesis guided by textual descriptions. We demonstrate the feasibility of our method in the context of SVG icons, giving rise to the name, \sysName. The key to success of \sysName~is to exploit the sequential nature of SVG:
An SVG script is composed of a sequence of paths, which in turn consists of a sequence of drawing commands (e.g., lines and curves, see Section \ref{method_tokenization}). We thus concatenate all SVG paths in a uniquely decodable way to form a command sequence. Since the text prompt is sequential in nature as well, it can be straightforwardly prepended to the command sequence.
We thus tokenize and mask the combined sequence in a way~\cite{aghajanyan2022cm3, fried2023incoder, bavarian2022efficient} that admits standard training of \sysName~(to predict the next token autoregressively), while enabling conditional SVG icon synthesis on the bidirectional context (also known as the filling-in-the-middle task). This can be effectively achieved by incorporating the right context into the left context (separated by a special token), which conforms to causal (i.e., autoregressive) masking.

We train \sysName~on a large-scale vector icon dataset \textit{FIGR-8-SVG}~\cite{clouatre2019figr}, consisting of monochromatic (i.e., black-and-white) SVG icons. We conduct comprehensive evaluations of \sysName~in terms of generation quality and diversity under different synthesis settings. Our experimental results show that \sysName~is superior to existing image-based and language-based methods in these two aspects both quantitatively and qualitatively. The synthesized SVG icons also show reasonable faithfulness to the corresponding text prompts. Moreover, we demonstrate the flexibility of \sysName~with multiple novel icon synthesis tasks, including icon editing, icon interpolation,  icon semantic combination, and icon design auto-suggestion.

\section{Related Work}
Our work is related to text-to-image generation (Section \ref{related_t2i}), vector graphics generation (Section \ref{related_vg}), and generative transformers (Section \ref{related_trans}).
Here we only provide a concise review of previous work that is closely related to ours, and a comprehensive treatment of the above areas is beyond the scope of this paper.

\subsection{Text-to-Image Generation}\label{related_t2i}
Generating images from text is a challenging task that has gained substantial attention in recent years, and has gone through three stages of development: Generative Adversarial Networks (GANs)~\cite{reed2016generative, zhang2017stackgan, zhang2018photographic, qiao2019mirrorgan, xu2018attngan,  kang2023scaling}, seq2seq models based on transformers~\cite{ramesh2021zero, ding2021cogview, yu2022scaling, ding2022cogview2, chang2023muse}, and diffusion models~\cite{nichol2021glide, saharia2022photorealistic, rombach2022high}.
Specifically, a GAN \cite{NIPS2014_5ca3e9b1} involves training two neural networks - a generator and a discriminator - to play a minmax zero-sum game. The system learns to generate new images by respecting the training data distribution and text conditions. Text-conditioned GANs~\cite{reed2016generative, zhang2017stackgan, zhang2018photographic, qiao2019mirrorgan, xu2018attngan,  kang2023scaling} are typically limited to modeling single and multiple object classes. Scaling them up to handle complex image datasets remains very challenging due to the instability occurred in the training procedure, until very recently~\cite{kang2023scaling}. Seq2seq models based on transformers operate by converting (and concatenating) the input text (and image) into a sequence of tokens for predicting another sequence of tokens that corresponds to the target image~\cite{ramesh2021zero, ding2021cogview, yu2022scaling, ding2022cogview2, chang2023muse}. Text-only and image-only self-attention and text-image cross-attention are canonical computational mechanisms in seq2seq models to capture intricate dependencies between text and image tokens.
Recently, diffusion models have emerged to be the new standard in text-to-image generation overnight. Typically, an unconditional diffusion model~\cite{ho2020denoising} initiates its process with Gaussian noise, and iteratively eliminates it to yield a natural image. Text-guided diffusion models~\cite{nichol2021glide, saharia2022photorealistic, rombach2022high} leverage the text embedding either as input or through cross-attention.
Although previous work tackles text-guided visual content generation like ours, they focus primarily on raster images with fixed resolution. In contrast, we aim for a different goal - text-guided vector icon synthesis with arbitrary scaling.

\subsection{Vector Graphics Generation} \label{related_vg}
In the early 2000s, SVG content can be created using PERL~\cite{probets2001vector} with plentiful drawing commands, but requires extensive human intervention. \citet{bergen2012automatic} automated the determination of the number and type of drawing commands by evolutionary computation to match a target raster image. These work can be seen as ancestors of vector graphics generation methods based on deep neural networks for learning editable SVG representations. SketchRNN~\cite{ha2017neural} is a pioneering deep representation learning model for vector sketches based on a seq2seq Variational Auto-Encoder (VAE)~\cite{kingma2013auto}. The encoder and the decoder were implemented by a bidirectional recurrent neural network (RNN) and an autoregressive RNN, respectively. The sketches were parameterized by polylines - a sequence of points with line segments drawn between consecutive points. \citet{lopes2019learned} incorporated a VAE-learned raster image representation to aid SVG font synthesis. The feasibility of the method is demonstrated only on glyphs with a maximum of $50$ commands. Modeling the layered structure of SVG, DeepSVG~\cite{carlier2020deepsvg} trained two transformer-based encoders in tandem to map SVG icons from commands to path-level representations,  and then to a global latent representation. Two decoders were paired up for SVG icon reconstruction. Although the reconstructed shapes look reasonable, DeepSVG fails to reproduce simple geometric relationships like perpendicularity and parallelism. Inspired by DeepSVG, \citet{aoki2022svg} made full use of the global latent representation in every stage of the decoder for synthesizing Chinese SVG characters.
Despite demonstrated success, the above-mentioned methods fail to support text-guided SVG generation.

One straightforward approach of text-guided SVG content synthesis is to first generate a raster image with a trained text-to-image generation model (e.g., DALL·E~\cite{ramesh2021zero} and Stable Diffusion~\cite{rombach2022high}), and then vectorize it using off-the-shelf image vectorization techniques (e.g., Potrace~\cite{selinger2003potrace} and LIVE~\cite{ma2022towards}). Another emerging line of research is to directly optimize SVG parameters for a pretrained vision-language model as the loss function. For instance, CLIPDraw~\cite{frans2021clipdraw} adopted the CLIP~\cite{radford2021learning} model as the optimization objective to measure the embedding distance between the vector-to-raster image and the input textual description. Apart from the CLIP distance, VectorFusion \cite{jain2022vectorfusion} leveraged the Score Distillation Sampling (SDS) loss~\cite{poole2022dreamfusion} based on a pixel-space text-to-image diffusion model. Despite different design philosophies, the vectorization-based and optimization-based methods suffer from similar limitations. First, the vision-language models as the key enablers are pretrained on raster images of complex natural scenes, and thus can hardly provide guidance in synthesizing SVG-style images with simple geometric primitives and flat colors. Second, the generated paths are often jagged and messy, failing to reproduce accurate geometric relations such as parallelism and perpendicularity. Third, the per-SVG optimization can be painfully slow, making it impractical for real-time applications. In contrast, our proposed system, \sysName, does not suffer from any of the above-mentioned problems. Once trained, \sysName~can perform text-guided vector icon synthesis automatically and efficiently.

\subsection{Transformers as Generative Models}\label{related_trans}
Owing to its inherent capability to capture long-term dependencies and support parallel computing, Transformers~\cite{vaswani2017attention} have emerged as a powerful class of generative models for producing a wide variety of outputs, ranging from natural languages~\cite{radford2019language, brown2020language, raffel2020exploring}, audios~\cite{huang2018music, li2019neural, valle2020flowtron}, and raster images~\cite{esser2021taming, chen2020generative}. Transformers can be made non-autoregressive and autoregressive. The non-autoregressive instantiation~\cite{Chang_2022_CVPR, ding2022cogview2, chang2023muse, zhang2021m6} suggests to leverage the bidirectional context using BERT-like~\cite{devlin2018bert} bidirectional Transformers for its sampling efficiency. The autoregressive instantiation~\cite{ramesh2021zero, yu2022scaling, ding2021cogview} emphasizes the importance of learning to predict the next token in a causal way. Together with scaling, it unlocks the emerging abilities of LLMs. Inspired by~\cite{aghajanyan2022cm3, bavarian2022efficient}, we unify non-autoregressive and autoregressive modeling of vector icons for various synthesis tasks.

\section{\sysName}

\begin{table*}[t]
\vspace{-2mm}
\centering
\small
\setlength{\tabcolsep}{0.25em}
\renewcommand{\arraystretch}{1.1}
\begin{tabular}{>{\centering\arraybackslash} p{2cm}>{\centering\arraybackslash} p{1cm} >{\centering\arraybackslash} p{2cm}>{\centering\arraybackslash} p{5cm}>{\centering\arraybackslash} p{4cm}}
    \toprule
    \textbf{Name} & \textbf{Symbol} & \textbf{Argument} & \textbf{Explanation} & \textbf{Example} \\
    \midrule
        Move To
        &
        \texttt{M}
        &
        \begin{tabular}{c}
             $x$, $y$
        \end{tabular}
        &
        \begin{tabular}{p{4.6cm}}
            Move the cursor to the specified point $(x, y)$.
        \end{tabular}
        &
        \begin{tabular}{c}
             \texttt{M} $\underline{20, 10}$\vspace{1.5mm}\\
             \includegraphics[width=1.5cm]{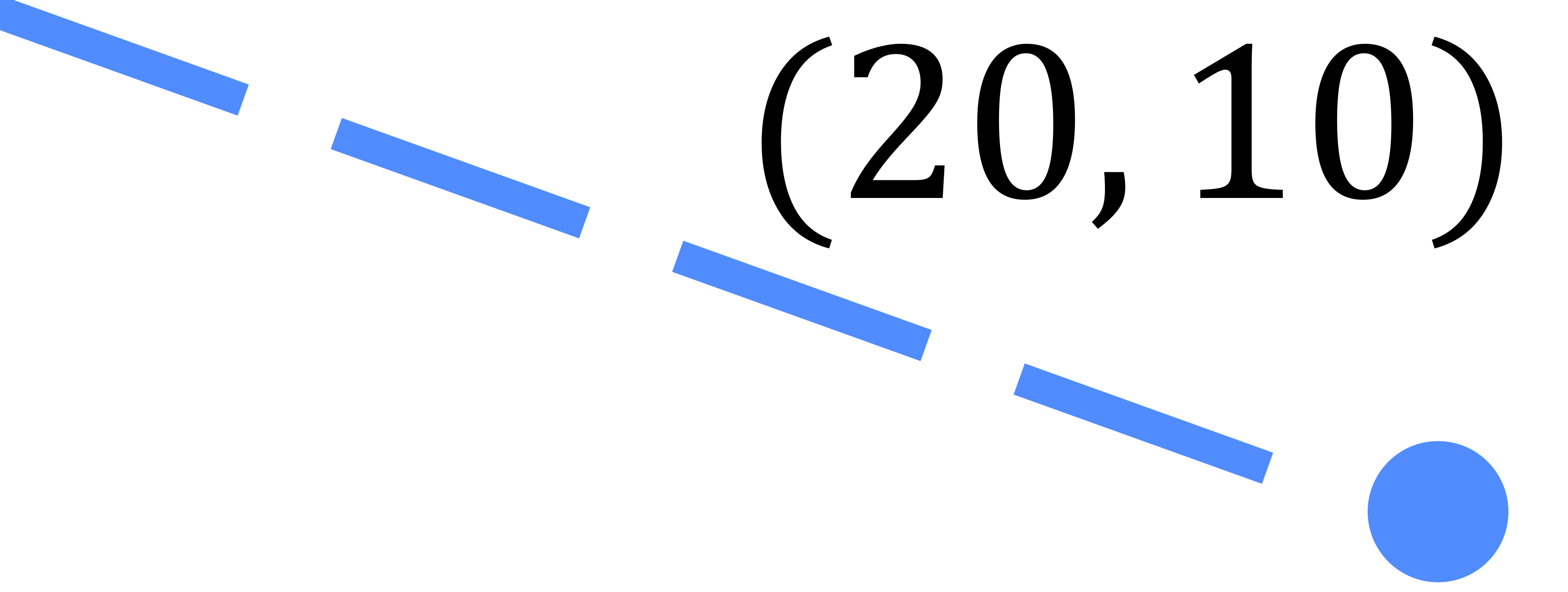}
        \end{tabular}
        \vspace{-1mm}
        \\
    \midrule
        Line To
        &
        \texttt{L}
        &
        \begin{tabular}{c}
             $x$, $y$
        \end{tabular}
        & 
        \begin{tabular}{p{4.6cm}}
        Draw a line segment from the current point to the specified point $(x, y)$.
        \end{tabular}
        &
        \begin{tabular}{c}
             \texttt{L} $\underline{20, 10}$\vspace{1.5mm}\\
             \includegraphics[width=1.5cm]{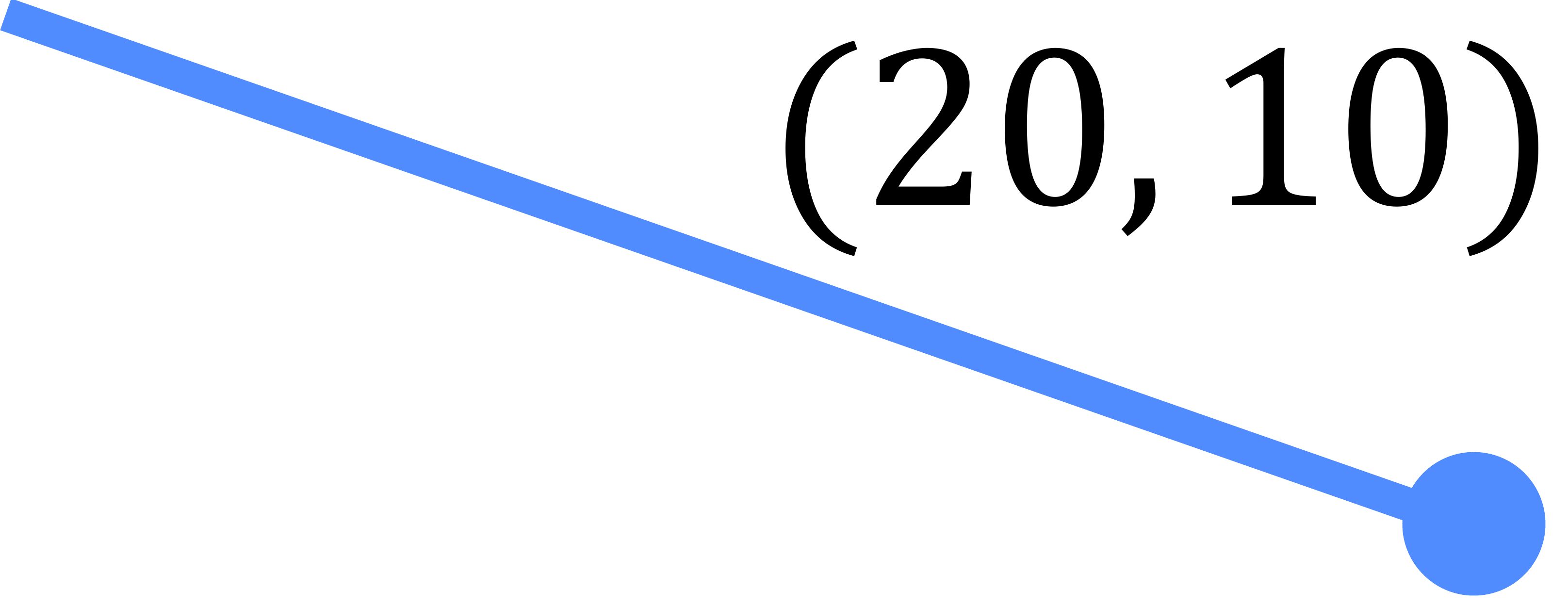}
        \end{tabular}
        \vspace{-1mm}
        \\
    \midrule
        Cubic B\'{e}zier
        &
        \texttt{C}
        &
        \begin{tabular}{c}
             $x_1$, $y_1$ \\ $x_2$, $y_2$ \\
             $x$, $y$
        \end{tabular}
        &
        \begin{tabular}{p{4.6cm}}
            Draw a curved path from the current point to the specified point $(x, y)$ using two control points $(x_1, y_1)$ and $(x_2, y_2)$.
        \end{tabular}
        &
        \begin{tabular}{c}
             \texttt{C} $\underline{13,2}\ \underline{7,8}\ \underline{20,10}$\vspace{1.5mm}\\
             \includegraphics[width=2cm]{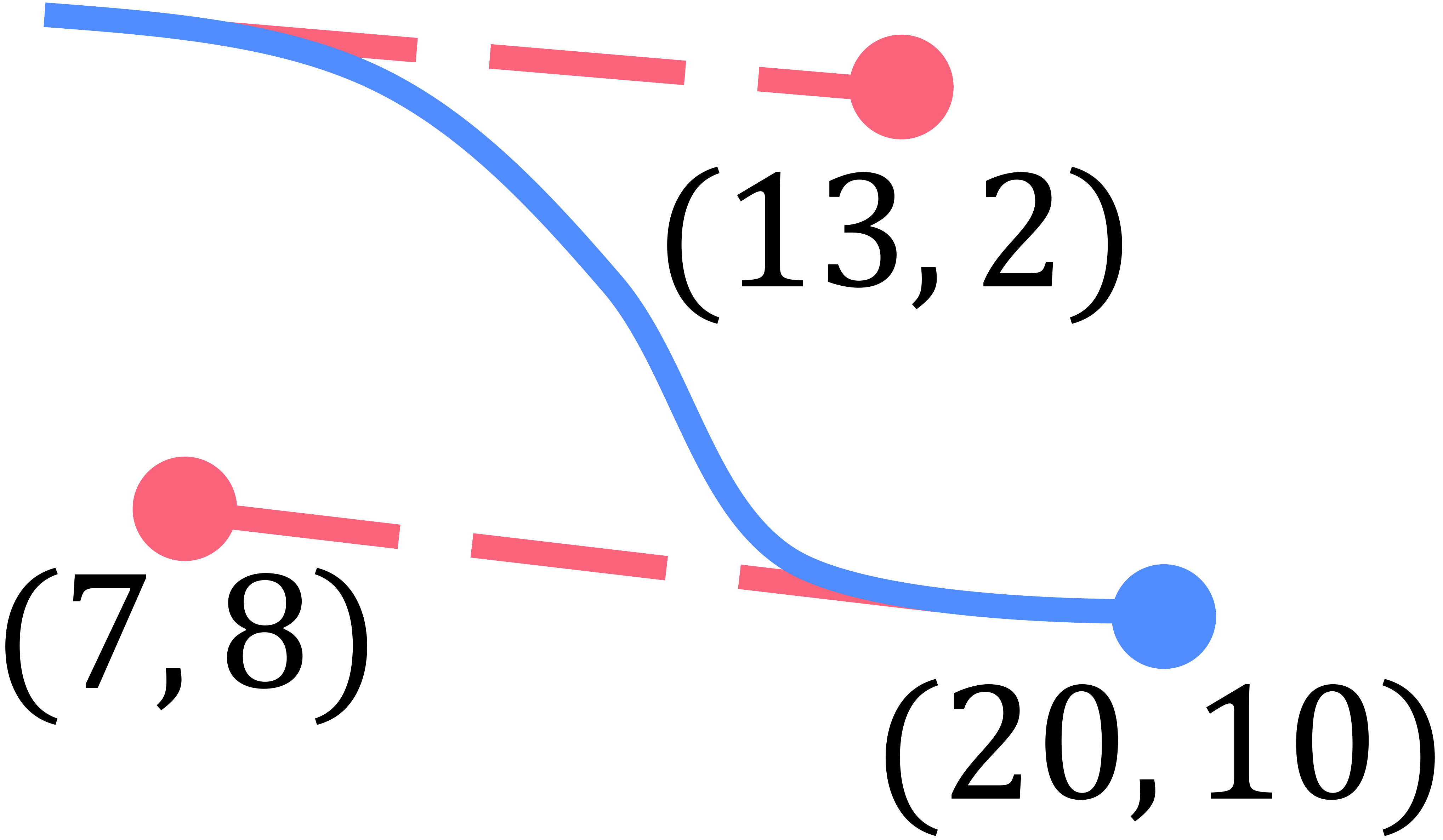}
        \end{tabular}
        \vspace{-1mm}
        \\
    \bottomrule
\end{tabular}
\captionsetup{width=0.81\textwidth}
\caption{Overview of SVG commands. In our implementation, we employ three simple yet expressive commands (namely, \texttt{M}, \texttt{L}, and \texttt{C}) to represent vector icons. In the Example column, we assume the current point is located at $(0,0)$. For a sequence of commands, the starting position of each command is the ending position of the previous command.}
\vspace{-4mm}
\label{tab:basic_commands}
\end{table*}

In this section, we first briefly introduce autoregressive models (Section \ref{method_ar}), and describe the SVG commands of vector icons, followed by our tokenization strategy (Section \ref{method_tokenization}). We then describe the "causal" masking strategy, which enables our autoregressive model to perform the filling-in-the-middle task (Section \ref{method_cm}). We next elaborate our model architecture (Section \ref{method_model}) and, finally, present the training objectives (Section \ref{method_training}).

\subsection{Preliminaries on Autoregressive Models}\label{method_ar}
An autoregressive model specifies that the current state depends only on its previous states. In probabilistic terms, this corresponds to the chain rule of probability:
\begin{equation}
p(S)=\prod_{n=1}^{N}p(S_n\vert S_1,\ldots, S_{n-1}),
\end{equation}
in which we factorize the joint probability of a sequence of random variables $ S = (S_1,\ldots, S_N)$ into a product of conditional probabilities. At the $n$-th instance, autoregressive  models take the values of previous $n-1$ random variables (or the most recent ones if a Markov window is applied) as input, and compute the conditional probability distribution of $S_n$, from which we are able to draw a sample as its prediction. Here we resort to autoregressive models for SVG icon synthesis because it fits naturally in the sequential nature of SVG and textural descriptions.

\subsection{SVG Representation and Tokenization}\label{method_tokenization}
SVG offers a range of features and syntax options, allowing users to create their original work with great flexibility. For example, the \texttt{Rect} command creates a rectangular shape controlled by the starting point, width, and height arguments, like \texttt{<Rect x="80" y="90" width="100" height="100"/>}. The \texttt{Transform} attribute applies an affine transformation to an existing shape, like \texttt{<Rect Transform="rotate (-10 50 100)" x="80" y="90" width="100" height="100"/>}. If we try to come up with a universal data structure to represent all possible SVG  commands and attributes, such a data structure would become highly complex, which may hinder the probabilistic modeling of SVG icons.
To bypass this issue, we choose to limit the number of commands and attributes, while maintaining their expressiveness to capture the essence of SVG icons. In other words, we seek a compact SVG representation that makes its probabilistic modeling easier.

Inspired by DeepSVG~\cite{carlier2020deepsvg}, we simplify every SVG icon by removing all attributes and using only three basic commands: {Move To, Line To, and Cubic B\'{e}zier} (see Table~\ref{tab:basic_commands} for explanations and examples). 
Other complex commands (e.g., \texttt{Rect}, \texttt{Circle}, and \verb|Ellipse|) can be approximated by combinations of these basic commands with negligible visual differences. For example, we can use four line segments to construct a \texttt{Rect}, and concatenate four B\'{e}zier curves to form a \texttt{Circle}.
An SVG text script $G$, after the simplification, contains $M$ \textit{paths}, $G=\{P_i\}_{i=1}^M$, where $P_i$ is the $i$-th path, and each path $P_i$ in turn consists of $N_i$ \textit{commands}, $P_i=\{C_i^j\}_{j=1}^{N_i}$, where $C_i^j$ is the $j$-th command in the $i$-th path. A command, $C_i^j=(U_i^j, V_i^j)$, contains its type $U_i^j\in \{\texttt{M}, \texttt{L}, \texttt{C}\}$ and the corresponding location argument $V_i^j$.

\begin{figure*}[t]
    \centering
    \vspace{-2mm}
    \includegraphics[width=0.9\textwidth]{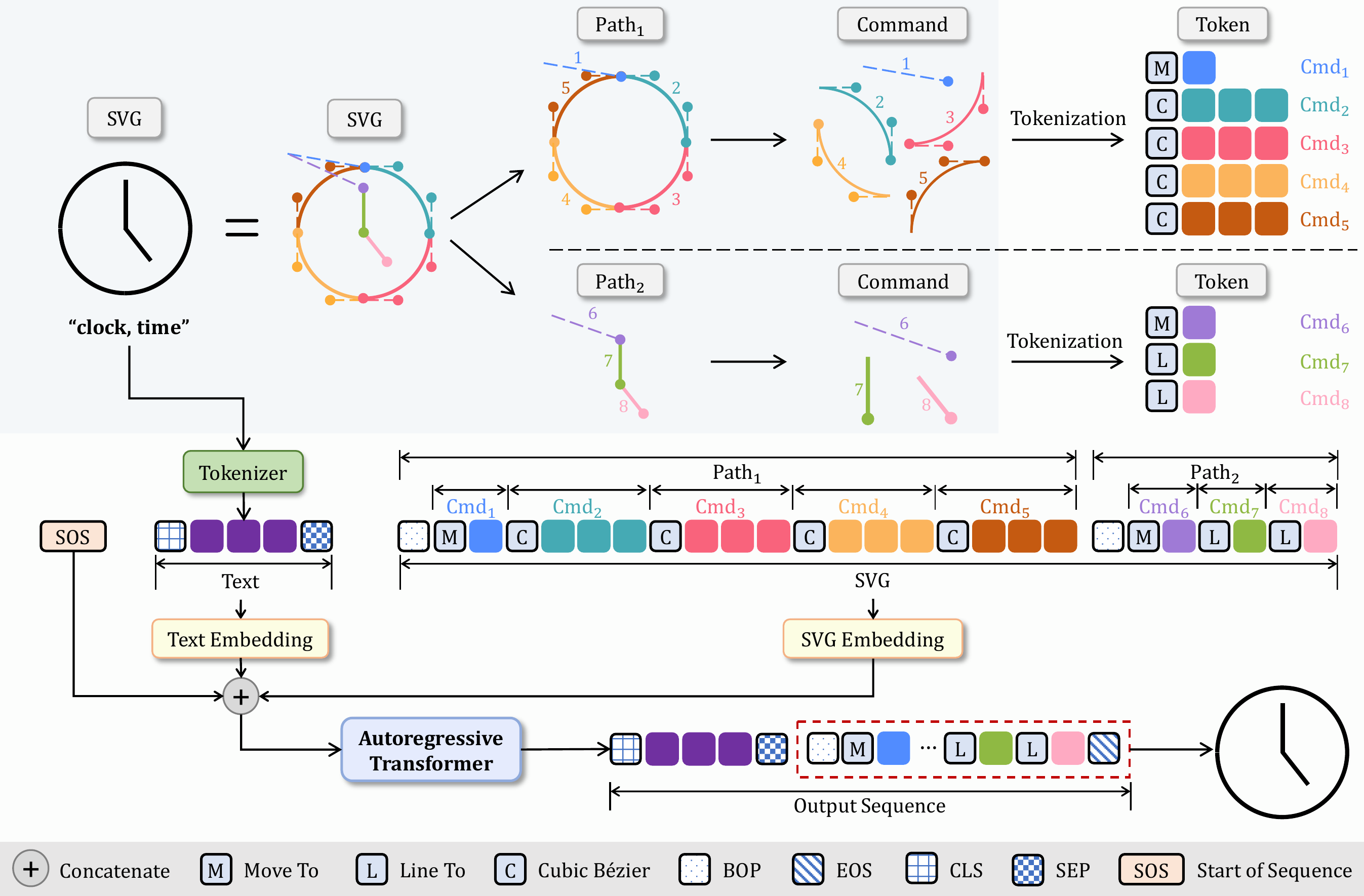}
    \vspace{-3mm}
    \caption{The system diagram of \sysName.
    In this example, the clock \underline{\textbf{icon}} has two paths, each comprising three types of basic commands (refer to Table~\ref{tab:basic_commands}). For $\text{Path}_1$, there is one \texttt{M} (Move To) command and four B\'{e}zier curves. For $\text{Path}_2$, there is one \texttt{M} command and two line segments. To tokenize these paths, we first concatenate the commands of the two paths to form a single sequence. We then convert each command's 2D location argument $(x, y)$ into 1D using the formula $x \times w + y$, where $w$ is the default SVG image width.
    We use a pretrained text encoder to tokenize and embed the \underline{\textbf{text}} "clock, time", where \texttt{<CLS>} and \texttt{<SEP>} mark the beginning and the end of the text input. We then concatenate the embedded text and SVG sequences, and add an \texttt{<SOS>} at the beginning. This concatenated sequence is sent to an autoregressive transformer for joint probability modeling.}
    \vspace{-3mm}
    \label{fig:pipeline}
\end{figure*}

The next step is to convert the SVG script into a discrete sequence of tokens, which are subject to autoregressive modeling. We introduce an intuitive SVG tokenization approach, which is composed of four major steps.
First, flatten the layered structure of the SVG script by concatenating
commands from different paths to form a single command sequence. To uniquely decode the flattened command sequence back to the layered representation, we prepend a special token, \texttt{<BOP>} (i.e., \textit{begin-of-path}), before the first command of each path.
Second, assign distinct tokens to each command type (i.e., \texttt{M}, \texttt{L}, \texttt{C}).
Third, map the 2D location argument to 1D using row-major order, which roughly halves the length of the token sequence. For example, suppose the default width of an SVG image is $w$, we transform a 2D location argument $(x, y)$ to a 1D argument using the formula $x\times w+y$.
Fourth, append a special token, \texttt{<EOS>} (i.e., \textit{end-of-SVG}), at the end of the sequence that indicates the completion of an SVG icon sequence. An example of a sequence created using the above steps is shown in Figure~\ref{fig:pipeline}.

\subsection{Masking Scheme} \label{method_cm}
Autoregressive models have been shown to be effective in generating token sequences from scratch, but they are constrained to doing so in a causal direction (i.e., from left to right). This constraint hinders SVG synthesis performance in tasks such as icon editing, where we need to fill in missing content based on the bidirectional context.
Some methods have been developed to expand the ability of autoregressive models to perform the filling-in-the-middle task by massaging training data without modifying the model architecture. For example, CM3~\cite{aghajanyan2022cm3} and InCoder~\cite{fried2023incoder} used a \textit{"causal" masking} strategy that randomly selects several chunks in the input sequence, and moves them to the end. In~\cite{bavarian2022efficient}, a similar strategy was implemented to learn to fill in the middle without sacrificing the original causal generation capability.
In training \sysName, we incorporate a similar "casual" masking strategy to unify non-autoregressive and autoregressive modeling of SVG token sequences.

For a given input sequence $S^{(0)}$, we first select a random chunk called \textit{span}, based on which we split the sequence into three parts, $[\texttt{Left}:\texttt{Span}:\texttt{Right}]$, where "$:$" represents concatenation. We replace the span with a special \texttt{<Mask>} token to obtain a new sequence $S^{(1)} = [\texttt{Left}:\texttt{<Mask>}:\texttt{Right}]$. Next, we add the same \texttt{<Mask>} token to the beginning of the span and add a \texttt{<EOM>} token (i.e., \textit{end-of-mask}) to the end of the span to create a new sequence $S^{(2)} = [\texttt{<Mask>}:\texttt{Span}:\texttt{<EOM>}]$. Finally, we concatenate $S^{(1)}$ and $S^{(2)}$ to form 
\begin{equation}\label{eq:maskeds}
    S = [\texttt{Left}:\texttt{<Mask>}:\texttt{Right}:\texttt{<Mask>}:\texttt{Span}: \texttt{<EOM>}],
\end{equation}
which is sent to the model for probabilistic modeling.
The masked sequence $S$ conveys the following information: 1) the first \texttt{<Mask>} token indicates the original position of the span, 2) the second \texttt{<Mask>} token denotes the beginning of the span, and 3) the \texttt{<EOM>} token marks the end of the span. During training, we randomly apply this masking strategy to $50\%$ of the training data, while leaving the remaining $50\%$ unchanged. We exclude the \texttt{<Mask>} token from the cross-entropy loss calculation to discourage its generation during inference.

We now explain how this masking technique enables autoregressive models to perform the filling-in-the-middle (i.e., non-autoregressive) generation without modifying the architecture. 
Suppose we have a token sequence $[\texttt{Left}:\texttt{Right}]$, and want to fill in the middle chunk between \texttt{Left} and \texttt{Right}. As in Eq.~\eqref{eq:maskeds}, we add two \texttt{<Mask>} tokens to the sequence to create$[\texttt{Left}:\texttt{<Mask>}:\texttt{Right}:\texttt{<Mask>}]$, and send it to the model for seq2seq generation until the \texttt{<EOM>} token occurs, giving rise to the sequence $[\texttt{Left}:\texttt{<Mask>}:\texttt{Right}:\texttt{<Mask>}:\texttt{Span}:\texttt{<EOM>}]$. After that, we move the predicted \texttt{Span} back to its original position, i.e., the position of the first \texttt{<Mask>} token, to obtain the final output $[\texttt{Left}:\texttt{Span}:\texttt{Right}]$. Since the model leverages both \texttt{Left} and \texttt{Right} contexts to fill in the middle \texttt{Span} chunk, we achieve non-autoregressive modeling through autoregressive prediction, and thus unify these two.

\subsection{Model Architecture} \label{method_model}
We employ the Transformer decoder~\cite{vaswani2017attention} to implement our autoregressive model, as it effectively captures the long-range interdependencies among various tokens that constitute a vector icon sequence. Specifically, the model consists of three modules: an SVG embedding module to encode the SVG sequence, a text embedding module to encode the text sequence, and a transformer (decoder) module to exploit text-SVG correlations, and learn the joint probability distribution of the combined token sequence, sampling from which produces novel sequences that are not present in the training set.

\noindent\textbf{SVG Embedding Module}.
As discussed previously, an SVG sequence is represented by six distinct categories of tokens:
1) Command type,
2) 1D location argument,
3) Begin-of-path token \texttt{<BOP>},
4) End-of-SVG token \texttt{<EOS>},
5) Mask token \texttt{<Mask>},
and 6) End-of-mask token \texttt{<EOM>}. By default, each icon is constrained within a $100\times 100$ bounding box, resulting in $100^2$ possible values for the 1D location argument. Thus, a one-hot vector $e$ with a dimension of $10,007(=3+100^2+1+1+1+1)$ suffices to represent all possible token cases. We then use a learnable embedding matrix ${W} \in \mathbb{R}^{D\times 10007}$ to transform the one-hot vector into an embedding vector of size $D$. We incorporate two extra learnable matrices ${W}^x, {W}^y \in \mathbb{R}^{D\times 100}$ to augment the location information as suggested in~\citep{xu2022skexgen}:
\begin{equation}
   {v}_i \leftarrow {W}{e}_i + {W}^x{e}_i^x + {W}^y{e}_i^y,
\end{equation}
where $e_i\in \mathbb{R}^{10007\times 1}$ is the one-hot vector of the $i$-th token, and $e_i^x, e_i^y\in \mathbb{R}^{100\times 1}$ are one-hot encoding of the 2D coordinates, respectively.

\noindent\textbf{Text Embedding Module}.
LLMs trained on a large corpus of textual data have the ability to capture intricate word interrelationships, including synonymy and antonymy, as suggested in~\citep{saharia2022photorealistic}. Here we make use of the tokenization and word embedding layers from a pretrained BERT~\cite{turc2019} model, and fix them to tokenize and embed textual inputs. The tokenizer adds a \texttt{<CLS>} token to the beginning of the text, and appends a \texttt{<SEP>} token at the end of the text, indicating the start and end of the text sequence, respectively.

\noindent\textbf{Transformer Module}. Our autoregressive transformer model consists of a stack of $12$ identical layers. Each layer is a standard transformer decoder block, comprising (masked) multi-head attention, layer normalization, and feed-forward layers, all interconnected via residual connections. The autoregressive transformer ultimately produces a $D$-dimensional vector at the $n$-th token position, which is conditioned on its preceding $n-1$ tokens. A linear layer followed by the softmax function is applied to obtain the probabilities of all possible tokens at the $n$-th position.

\subsection{Training Objective} \label{method_training}
Since the text descriptions and SVG scripts in the training set have varying lengths, we pad both the text token sequence $S^{\mathrm{text}}$ and the icon token sequence $S^{\mathrm{icon}}$ with zeros to a fixed length ($50$ for the text and $512$ for the icon in our implementation). We then concatenate them to obtain the target sequence $S=[S^{\mathrm{text}}:S^{\mathrm{icon}}]$.

The autoregressive transformer is trained to predict the next token based on previous tokens. To prepare the input sequence $S^{\mathrm{in}}$, we remove the last token of $S$ (i.e., \texttt{<EOS>}), and add an \texttt{<SOS>} token at the start. This essentially shifts $S$ to the right by one position, which enables initial autoregressive prediction with an empty context. The transformer outputs a token sequence $\hat{S}=[\hat{S}^{\mathrm{text}}:\hat{S}^{\mathrm{icon}}]$ sequentially. Our objective is to minimize the cross-entropy loss between the target and output tokens at each position, and then combine text and icon losses with a weighted sum~\cite{ramesh2021zero}:
\begin{equation}
  \begin{aligned}
    \ell^{\mathrm{text}} &= \mathrm{CE}(S^{\mathrm{text}}, \hat{S}^{\mathrm{text}}),\\
    \ell^{\mathrm{icon}} &= \mathrm{CE}(S^{\mathrm{icon}}, \hat{S}^{\mathrm{icon}}),\\
    \ell^{\mathrm{total}} &= \ell^{\mathrm{text}} + \lambda\ell^{\mathrm{icon}},
  \end{aligned}
\end{equation}
where $\mathrm{CE}()$ is the standard cross-entropy function, and $\lambda = 7.0$ is the weighting to control the relative importance between the text and icon reconstruction.

\section{Experiments}
In this section, we first present in detail the data processing procedure for the icon dataset with textual descriptions (Section~\ref{experiment_setup}). We then introduce an ablation study to validate the efficacy of the model architecture under both conditional and unconditional generation settings (Section~\ref{experiment_ablation}). Lastly, we compare the proposed \sysName~with alternative solutions, demonstrating that \sysName~yields higher-quality results (Section~\ref{experiment_comparison}). 

\subsection{Data Preparation}
\label{experiment_setup}

\begin{figure}[t]
\centering
    \vspace{-2mm}
    \includegraphics[width=7.7cm]{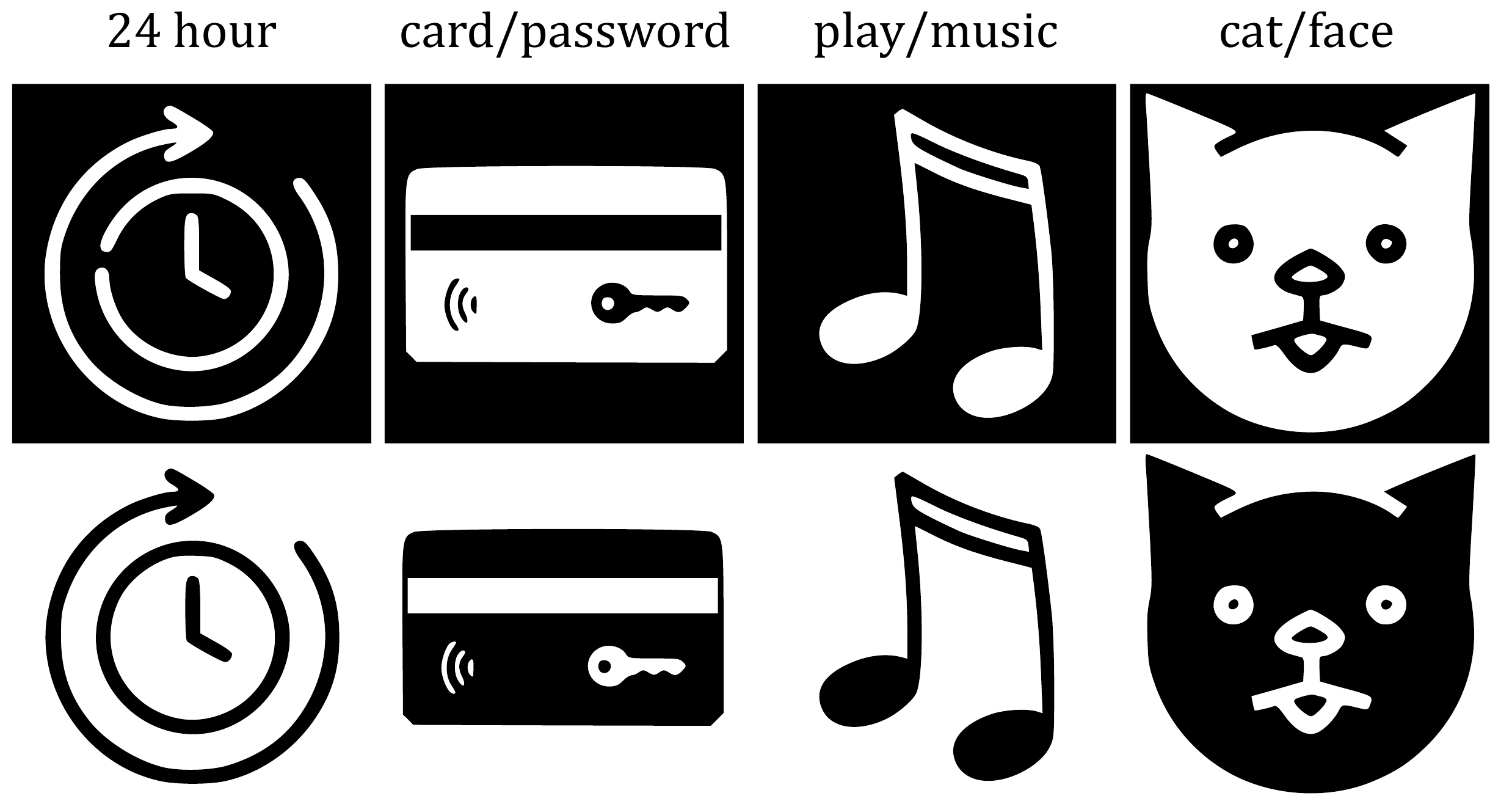}
    \vspace{-3mm}
    \caption{Monochromatic icon samples from the \textit{FIGR-8-SVG} dataset (1st row). Each icon is associated with several discrete keywords as textual descriptions. We remove the black box to improve the visual quality (2nd row).}
    \vspace{-3mm}
\label{fig:svg_dataset}
\end{figure}

\textbf{SVG Dataset}. We use the \textit{FIGR-8-SVG} dataset~\cite{clouatre2019figr} that consists of $1.5$ million monochromatic (black-and-white) vector icons. Typically, the first step for SVG data processing is transforming icons with varied grammars into standardized representations. Fortunately, in the \textit{FIGR-8-SVG} dataset, all icons have been converted to a uniform representation with discretized arguments. We show several icon examples from the dataset in the first row of Figure~\ref{fig:svg_dataset}. We further enhance the visual attractiveness of each icon by removing the outer black box. The corresponding enhanced icons are shown in the second row.
After the SVG tokenization described in Section~\ref{method_tokenization}, we set the maximum length of an icon sequence to $512$, filtering out those with longer lengths. This results in about $1.1$ million samples, among which, we select $300,000$ samples for model training and experimentation at the current stage. We partition these samples into $90\%$ for training, $5\%$ for validation, and $5\%$ for testing.

\noindent\textbf{Text Input}.
In the \textit{FIGR-8-SVG} dataset, every vector icon is annotated with discrete keywords, such as "\texttt{cat/face}". Training \sysName~only with keywords would constrain its capacity to generate icons from natural language phrases and sentences.
Inspired by InstructPix2Pix~\cite{brooks2022instructpix2pix}, which fine-tunes GPT-3~\cite{brown2020language} to produce editing instructions and captions, we use LLMs (ChatGPT\footnote{\url{https://openai.com/blog/chatgpt}} in particular) to expand these keywords into natural language phrases and sentences. The prompt given to ChatGPT is "\texttt{Write the simplest sentence from keywords: \#\{keywords\}. Do not add additional facts}".

\noindent\textbf{Implementation Details}. We implement \sysName~using \textit{PyTorch}. The training process employs the Adam optimizer~\cite{kingma2014adam} with a learning rate of $0.0006$, along with linear warm-up, decay, and gradient clipping. The dropout rate is set to $0.1$. Shuffled discrete keywords, natural language phrases and sequences (generated by ChatGPT), and blank text are, respectively, trained with ratios of $60\%$, $30\%$, and $10\%$, with a total minibatch size of $192$. We train \sysName~for $300$ epochs. We use an NVIDIA A100 GPU in the following experiments to test \sysName,  which takes $1.38$ seconds to generate an SVG icon sequence on average.

\begin{figure}[t]
\centering
    \vspace{-2mm}
    \includegraphics[width=7.7cm]{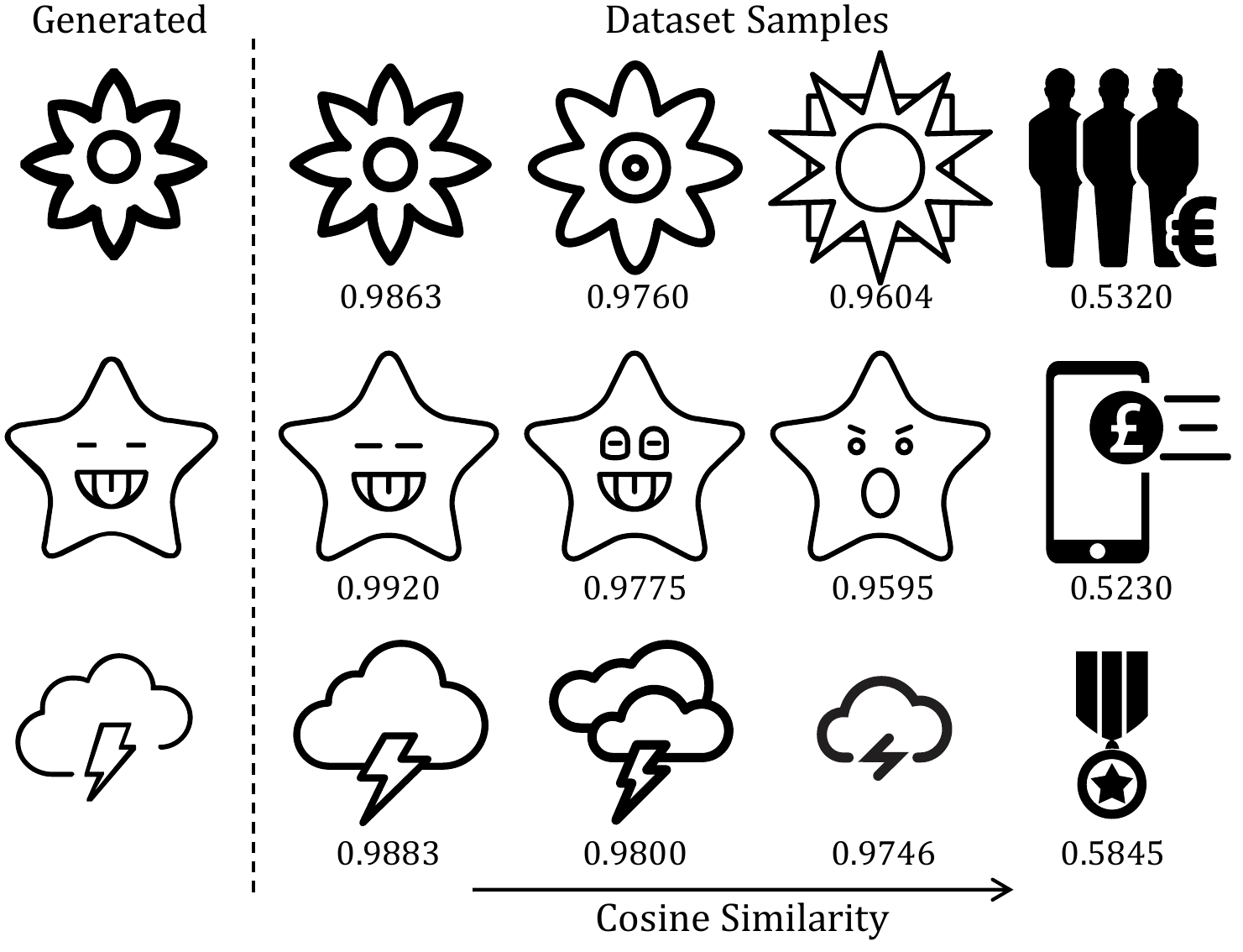}
    \vspace{-3mm}
    \caption{We use the CLIP image encoder to extract image features, and calculate the cosine similarity between the generated icons and the samples in the dataset. As visually distinct icons may have high cosine similarity scores due to the simple black strokes against the white background, we set a relatively high threshold of $0.98$ to determine whether two icons are identical when computing the "Uniqueness" and "Novelty" scores.}
    \vspace{-3mm}
\label{fig:cosine_similarity}
\end{figure}

\noindent\textbf{Metrics}.
To evaluate the quality of the generated SVG icons, we use the Fréchet Inception Distance (FID)~\cite{heusel2017gans}, which measures the distance between the image features of synthesized and "ground-truth" SVG icons from \textit{FIGR-8-SVG}. Specifically, we obtain the image features of rendered and rasterized SVG icons using the CLIP image encoder~\cite{radford2021learning}.
We also compute the CLIP score to measure text-SVG alignment {(i.e., the semantic similarity between the text input and the visual icon output)}. 
Additionally, following SkexGen~\cite{xu2022skexgen}, we calculate the "Novelty" and the "Uniqueness" scores, which refer to the proportion of generated data absent from a target set (i.e., the training set from \textit{FiGR-8-SVG}) and occurring only once among all generated samples, respectively.
We generate $20,000$ icons unconditionally and $7,000$ icons with textual guidance. The cosine similarity between the CLIP features is used to determine whether the two icons are identical with a threshold of $0.98$. Figure~\ref{fig:cosine_similarity} gives an intuitive visual comparison of different similarity values.

\begin{figure*}[t]
    \centering
    \vspace{-2mm}
    \includegraphics[width=0.95\textwidth]{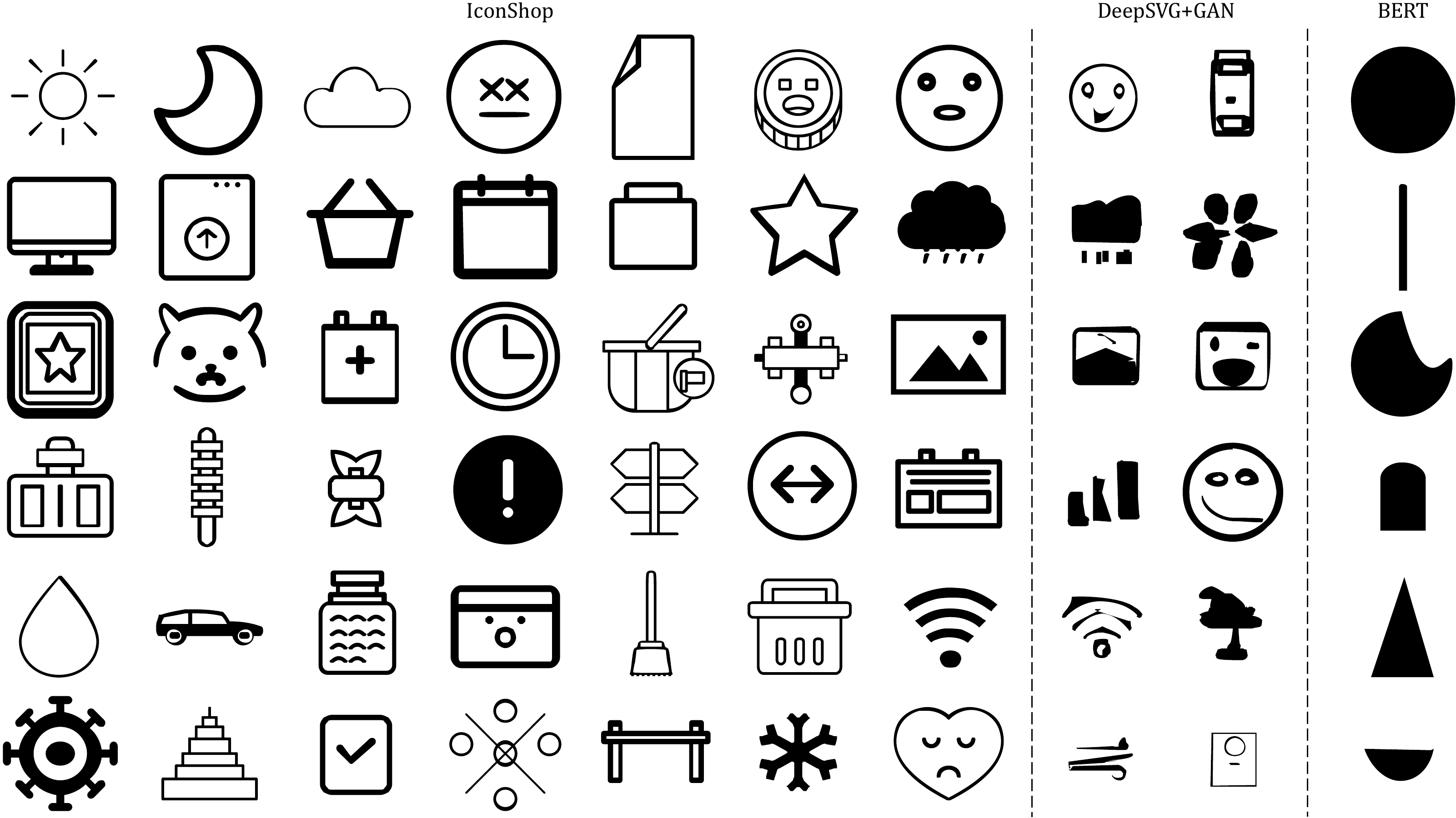}
    \vspace{-3mm}
    \caption{Icons randomly generated by \sysName, DeepSVG+GAN, and BERT, respectively. Our approach creates icons with form consistency, high-precision of recognizability, geometry simplicity, and good composition.  Icons produced by DeepSVG+GAN do not meet such desired properties, while BERT only synthesizes basic geometric shapes with essentially no semantics.}
    \vspace{-3mm}
    \label{fig:random_generation}
\end{figure*}

\begin{figure*}[t]
    \centering
    \vspace{-2mm}
    \includegraphics[width=0.95\textwidth]{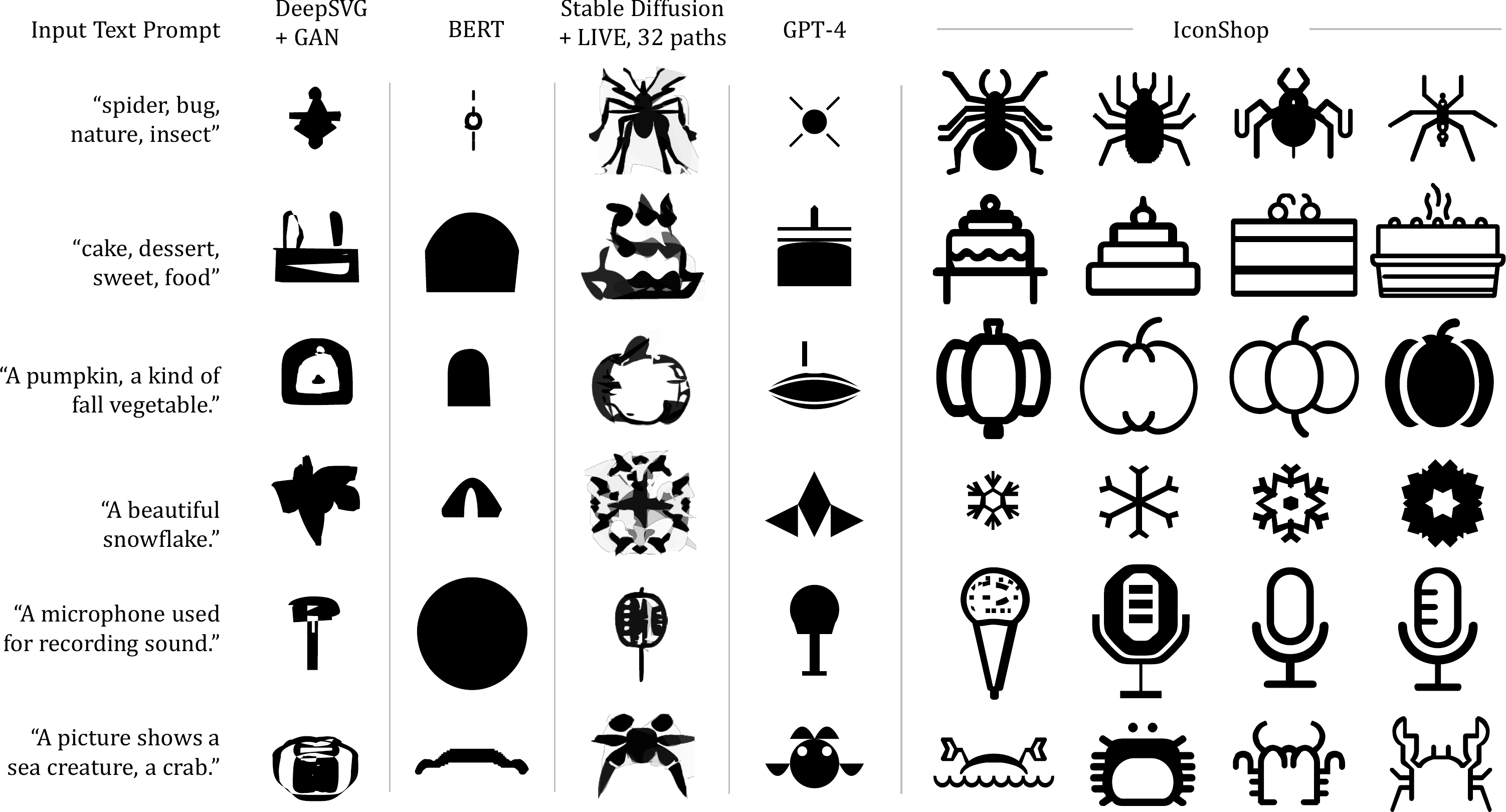}
    \vspace{-3mm}
    \caption{We evaluate the text-guided icon synthesis of \sysName~against several other methods, including DeepSVG+GAN, BERT, the image-based method (Stable Diffusion+LIVE), and the language-based method (GPT-4). The results show that Stable Diffusion+LIVE provides poor vectorization results, in terms of geometrical and semantic precision of icon forms, while GPT-4 is only able to combine basic geometric shapes with limited recognizability. Our \sysName~clearly outperforms the competing methods in terms of icon generation quality.}
    \vspace{-3mm}
    \label{fig:text_guided_generation}
\end{figure*}

\subsection{Ablation on Different Network Architecture}\label{experiment_ablation}
In this subsection, we conduct two ablation studies to illustrate the effects of two key components of \sysName~on the synthesis performance. The first study shows the importance of exploiting the sequential nature of the SVG icons in generation by comparing our seq2seq model to a GAN model. The second study demonstrates the necessity of our autoregressive model augmented with the "causal" masking scheme by comparing it to a non-autoregressive training strategy (i.e., BERT~\cite{devlin2018bert}).

\noindent\textbf{Seq2seq versus Layered Modeling}.
As previously mentioned, an SVG file is layered in structure, consisting of higher-layer paths and lower-layer commands.
Here we use DeepSVG+GAN as a baseline to testify our choice of the seq2seq generation mechanism.

The original DeepSVG reconstructs an SVG icon by first obtaining path-level representations and then aggregating them to a global representation. To make a fair comparison between \sysName~and DeepSVG, we need to enable the original DeepSVG to generate new icons and support text-guided generation. To achieve these goals, we retrain DeepSVG for SVG icon reconstruction on our dataset. We then train a conditional GAN model, which takes the textual feature as input and predicts the latent representation precomputed from the encoder of DeepSVG as output, which can be straightforwardly decoded into an SVG icon by passing through the decoder of DeepSVG. In addition to text-guided generation, we also enable the GAN to generate icons unconditionally by replacing the text feature with random noise $10\%$ of the time during training.

We compare both \emph{random generation} and \emph{text-guided generation} performance.
For random generation, we produce $20,000$ icons for each method. For text-guided generation, we choose both discrete keywords and natural phrases/sentences commonly used in design scenarios as text input and generate $7,000$ icons for each model. As shown in Figures~\ref{fig:random_generation} and \ref{fig:text_guided_generation}, \sysName~is capable of generating visually recognizable and eye-catching icons while preserving salient geometric relationships, such as perpendicularity, parallelism, and symmetry. We believe such consistent high-quality  generation results arise because we prioritize the sequential modeling of SVG icons. In contrast, DeepSVG+GAN presents visually worse results than ours, which we attribute to its averaging operation over commands and paths, resulting in a loss of geometric details in the generated icons. Please refer to the project page for more qualitative results.

We also quantitatively evaluate the random generation and text-guided generation results in Tables \ref{tab_random_generation} and \ref{tab_text_guided}. We find that \sysName~synthesizes results with significantly lower FID scores in both random and text-guided generation tasks, which provides a strong indication of the superior synthesis capability of \sysName. Regarding Uniqueness and Novelty as measures of generation diversity, \sysName~performs comparably well to DeepSVG+GAN. Nevertheless, it is important to note that the high Uniqueness and Novelty values achieved by DeepSVG+GAN are largely attributed to the noticeable visual distortions (e.g., jittering curves) it suffers from rather than novel instantiations of the same object/concept, as evident in Figure~\ref{fig:random_generation}. We regard this as "fake diversity" if it is solely interpreted without being conditioned on acceptable generation quality. Moreover, \sysName~obtains the highest CLIP score in the text-guided generation task, signaling the outstanding ability to produce icons that accurately reflect the text semantics.

\begin{table}[t]
    \begin{subtable}[h]{\linewidth}
        \footnotesize
        \centering
        \begin{tabular}{cccc}
            \hline
            & FID $\downarrow$ & Uniqueness$\% \uparrow$ & Novelty$\% \uparrow$ \\
            \hline
            DeepSVG+GAN & $11.95$ & $\mathbf{98.72}$ & $\mathbf{99.22}$ \\
            BERT & $43.61$ & $2.06$ & $19.90$ \\
            IconShop & $\mathbf{6.08}$ & $78.77$ & $85.10$ \\
            \hline
        \end{tabular}
        \caption{Random Generation}
        \label{tab_random_generation}
    \end{subtable}
    \hfill
    \begin{subtable}[h]{\linewidth}
        \footnotesize
        \centering
        \begin{tabular}{ccccc}
            \hline
            & FID $\downarrow$ & Uniqueness$\% \uparrow$ & Novelty$\% \uparrow$ & CLIP Score $\uparrow$ \\
            \hline
            DeepSVG+GAN & $12.01$ & $\mathbf{97.59}$ & $\mathbf{99.01}$ & $21.78$ \\
            BERT & $35.10$ & $14.41$ & $50.30$ & $22.03$ \\
            IconShop & $\mathbf{4.65}$ & $68.29$ & $68.60$ & $\mathbf{25.74}$ \\
            \hline
        \end{tabular}
        \caption{Text-Guided Generation}
        \label{tab_text_guided}
    \end{subtable}
    \vspace{-3mm}
    \caption{We evaluate \sysName~through random generation and text-guided generation tasks. We use the FID score, with features extracted from the CLIP image encoder, to assess generation quality. We also compute the percentage of unique and novel icons as measures of generation diversity. For text-guided generation, we employ the CLIP score to show the semantic alignment between the text input and the generated icons.}
    \vspace{-3mm}
    \label{tab:temps}
\end{table}

\noindent\textbf{Autoregressive versus Non-Autoregressive Modeling}.
Recent research~\cite{Chang_2022_CVPR, chang2023muse} shows that a non-autoregressive training paradigm for generating raster images is feasible. 
These approaches use bidirectional Transformer{s} (i.e., BERT) as the fundamental model architecture. The departure from the autoregressive to non-autoregressive modeling allows parallel prediction of multiple tokens, and opens the door to diverse image editing tasks such as inpainting. We explore the possibility of using  BERT to produce SVG token sequences, as an alternative to our "causal" masking scheme. It is noteworthy that, unlike raster image generation which is often fixed in length, SVG icon generation expects output sequences of varying lengths as a manifestation of generation diversity. This makes it  a more difficult task because in SVG icon generation, BERT needs to not only model the likelihood of next tokens but also determine where to terminate the sequence. 

After training the BERT model on the same dataset as ours, we assess the generation quality and diversity of its outputs.
Quantitative evaluations are conducted similarly to DeepSVG+GAN in both random (Table \ref{tab_random_generation}) and text-guided generation (Table \ref{tab_text_guided}). 
BERT produces worse results in both tasks than ours and DeepSVG+GAN.
The qualitative results in Figure~\ref{fig:random_generation} also suggest that BERT can only generate simple geometric shapes such as circles and rectangles that are  relatively insipid. A closer inspection of the filled tokens reveals that the end-of-SVG token, \texttt{<EOS>}, may frequently appear in multiple positions due to parallel prediction, which results in early termination when reconstructing SVG icons.
Therefore, despite the impressive ability of the non-autoregressive BERT in sequence editing, it is inferior to the autoregressive counterpart (as in \sysName) in modeling and producing sequences of varying lengths.

\subsection{Comparison to the State-of-the-Art}\label{experiment_comparison}
We compare the proposed \sysName~to two types of text-to-SVG generation schemes: image-based and language-based methods. For the former, we use Stable Diffusion~\cite{rombach2022high} to create raster images, which are transformed into SVG images using the LIVE~\cite{ma2022towards} program. To encourage icon-style results, we include keywords (like "monochrome" and "line art") in the text prompts. For the latter, we employ GPT-4, arguably the best-performing LLM so far, to generate SVG scripts directly. To initiate the conversation, we provide a text prompt at the system level asking it to act as an SVG code generator. 

Figure~\ref{fig:text_guided_generation} shows icons synthesized by the competing methods. We find that the outcomes of the Stable Diffusion model often fall short of expectations. After all, it is trained on raster images, which as expected struggles to produce icon-style images even with some prompt engineering.
After applying LIVE, the resulting SVG icons usually present unsmooth and inconsistent structures, many of which are not semantically recognizable. In addition,
Stable Diffusion is inefficient in generation due to the iterative optimization-based vectorization by LIVE. With respect to the language-based method, GPT-4 has relatively strong capabilities in generating SVG icon scripts purely from text prompts, with moderate text-SVG alignment. Nevertheless, the results are manifested as simple combinations of primitive shapes with no complex overlaying, which are inadequate for graphics design, and also suffer from the recognizability problem (see the third and fifth rows). In stark contrast, our \sysName~produces results with the highest visual quality in terms of form consistency, high precision of recognizability and text-SVG alignment, and geometry simplicity. Reasonable generation diversity is also well observed from \sysName. More visual results of the text-guided generation can be found in our project page.

\begin{figure*}[t]
    \centering
    \vspace{-2mm}
    \includegraphics[width=0.8\textwidth]{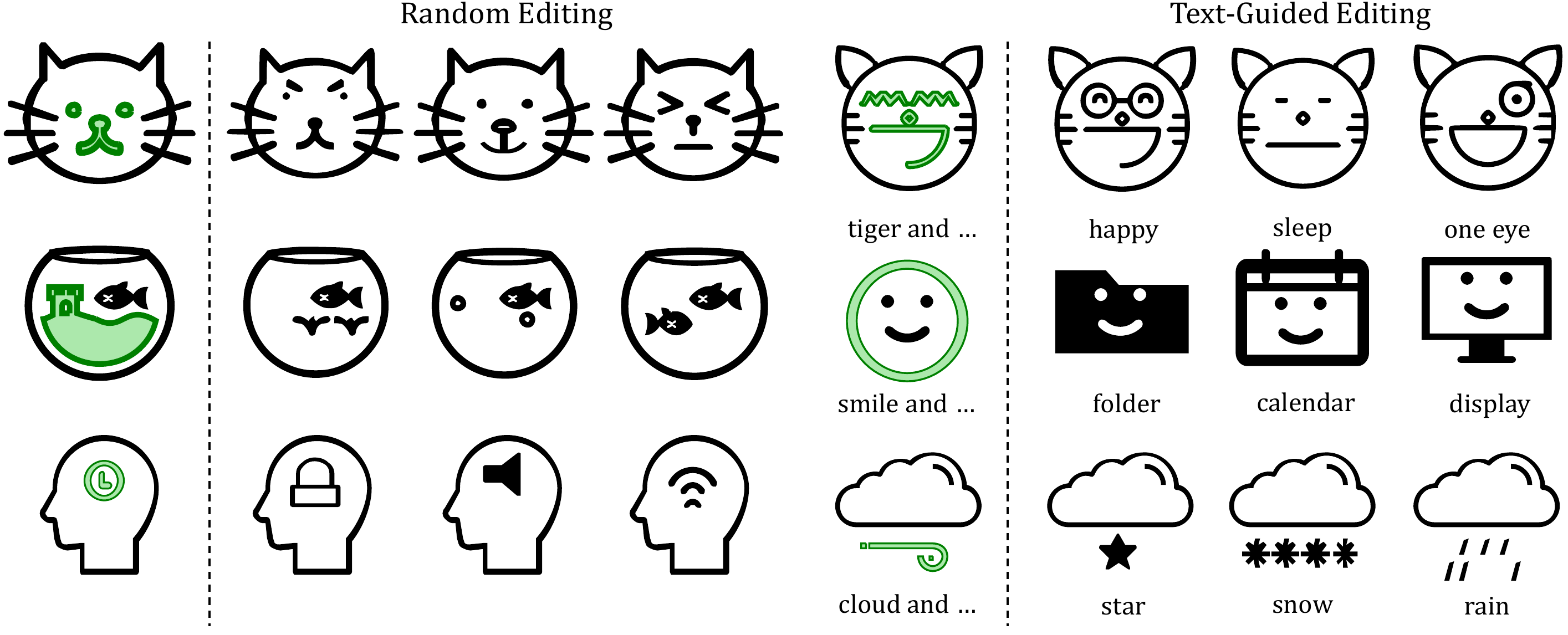}
    \vspace{-3mm}
    \caption{\sysName~enables both random and text-guided icon editing. We use the green color to highlight the paths to be edited.}
    \vspace{-3mm}
    \label{fig:editing}
\end{figure*}

\subsection{Subjective User Study}

\begin{table}[b]
\footnotesize
\begin{tabular}{lccccc}
\toprule
\multicolumn{1}{l}{} & \multicolumn{5}{c}{User Selection\% $\uparrow$} \\ \cline{2-6} 
& \begin{tabular}[c]{@{}c@{}}DeepSVG\\ + GAN\end{tabular} & \begin{tabular}[c]{@{}c@{}}Stable Diffusion\\ + LIVE\end{tabular} & GPT-4 & IconShop & Dataset \\ \hline
Quality (random) & $54.09$ & $15.95$ & $2.95$ & $82.11$ & $\mathbf{83.71}$ \\
Quality (text)   & $51.90$ & $49.49$ & $2.15$ & $\mathbf{96.33}$ & \textbackslash{} \\
Alignment (text)   & $29.24$ & $72.78$ & $1.77$ & $\mathbf{96.20}$ & \textbackslash{} \\
\bottomrule
\end{tabular}
\caption{Subjective user study results. In each task, we report the average percentage of selected (i.e., high-quality) icons by the users.}
\vspace{-5mm}
\label{tab:user_study}
\end{table}

To formally validate the perceptual gains by \sysName, we conduct a subjective user study, consisting of three tasks to visually assess 1) random generation quality, 2) text-guided generation quality, and 3) text-SVG alignment.
In the first task, we first familiarize users with high-quality icons from the training set. We then present them a total of $15\times 5$ icons in random order, one-fifth of which are generated by (or sampled from) DeepSVG+GAN, Stable Diffusion+LIVE, GPT-4, IconShop, and \textit{FIGR-8-SVG}, respectively. Users are forced to give a binary decision of whether each presented icon is of high quality. In the second task, we randomly select ten text prompts, and generate the corresponding icons by the four competing methods. Users are then asked to select two icons that they believe are of the best visual quality. The setup of the third task is identical to that of the second task. The difference is that this time users need to pick two icons that best match the corresponding text prompt.

We conduct the user study via an online questionnaire, with $79$ users participating in the study. For the first task, we obtain $79$ (users) $\times$ $15$ (icons) $\times$ $5$ (methods) = $5,925$ human judgments. We calculate the average percentage of high-quality icons identified by users for each method, and list the results in the first row of Table~\ref{tab:user_study}. We see that \sysName~approaches the performance to that of  the "Dataset" as the upper bound, and is clearly better than the other three methods. This shows that \sysName~is able to produce high-quality icons that consistently "fool" the subjects in the random (unconditional) generation setting. 

For the second and third tasks, we, respectively, collect $79$ (users) $\times$ $10$ (text prompts) $\times$ $2$ (selections) = $1,580$ human judgments for each task. We report the average percentage of user-selected icons by each method in the last two rows of Table~\ref{tab:user_study}. It is clear that icons synthesized by \sysName~are most frequently selected, indicating the highest quality and the best text-SVG alignment in the text-guided generation task. We also perform  one-way  ANalysis Of VAriance (ANOVA) tests, and  the $p$-values of the three tasks are all less than $0.001$, suggesting that the perceptual gains by \sysName~are statistically significant.
In summary, the proposed \sysName~demonstrates the highest quality for both random and text-guided generation, with a strong text-SVG alignment.

\section{Application}
In this section, we explore four practical applications of \sysName: icon editing, icon interpolation, icon semantic combination, and icon design auto-suggestion. These applications streamline the process of vector icon synthesis, enhancing user productivity and experience significantly.

\begin{figure*}[t]
    \centering
    \includegraphics[width=0.9\textwidth]{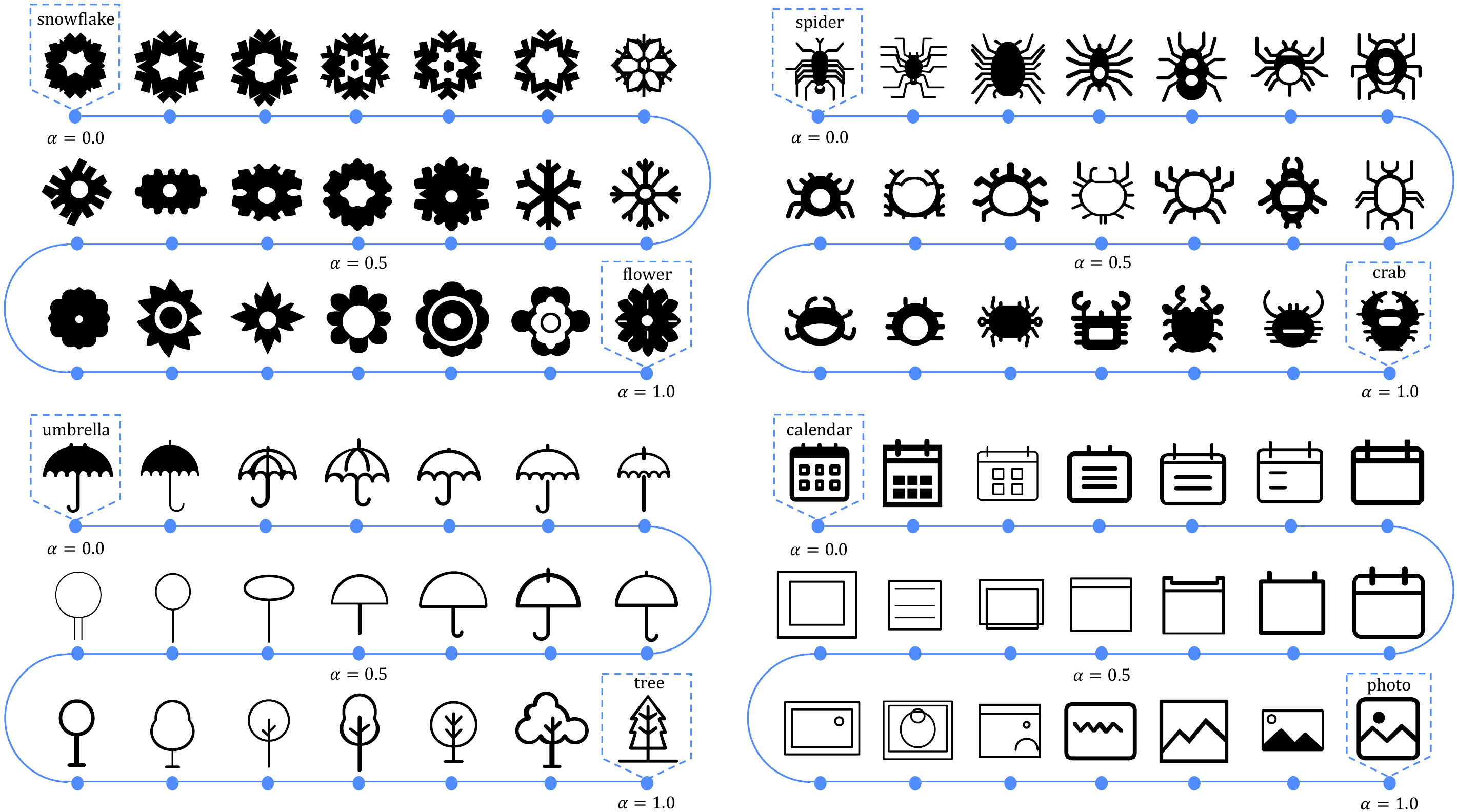}
    \vspace{-3mm}
    \caption{We linearly interpolate two text embedding vectors using the formula $v = (1 - \alpha) \cdot v^{(1)} +\alpha \cdot v^{(2)}$, where $\alpha \in [0, 1]$. Subsequently, we generate icons corresponding to each interpolated vector $v$. The results show that \sysName~learns a smooth mapping between the text and SVG spaces.}
    \vspace{-3mm}
    \label{fig:interpolation}
\end{figure*}

\subsection{Icon Editing}
Thanks to the unification of non-autoregressive and autoregressive modeling through the "causal" masking strategy outlined in Section~\ref{method_cm}, the proposed \sysName~facilitates icon editing as exemplified in Figure~\ref{fig:editing}. \sysName~is capable of filling in missing content based on the bidirectional context, in either random or text-guided generation scenario. This leads to precise, consistent, and diverse restoration of the missing paths in icons.

\subsection{Icon Interpolation}
Text-to-image models demonstrate an impressive ability to combine textual inputs, encouraging generation of novel concepts that do not exist in the training data, such as "\texttt{avocado chair}" generated by DALL·E~\cite{ramesh2021zero}. In our experiments, we find that \sysName~also learns to create innovative and reasonable combinations, as shown in Figure~\ref{fig:text_combination}.

\subsection{Icon Semantic Combination}
\begin{figure}[t]
\centering
    \vspace{-2mm}
    \includegraphics[width=6.5cm]{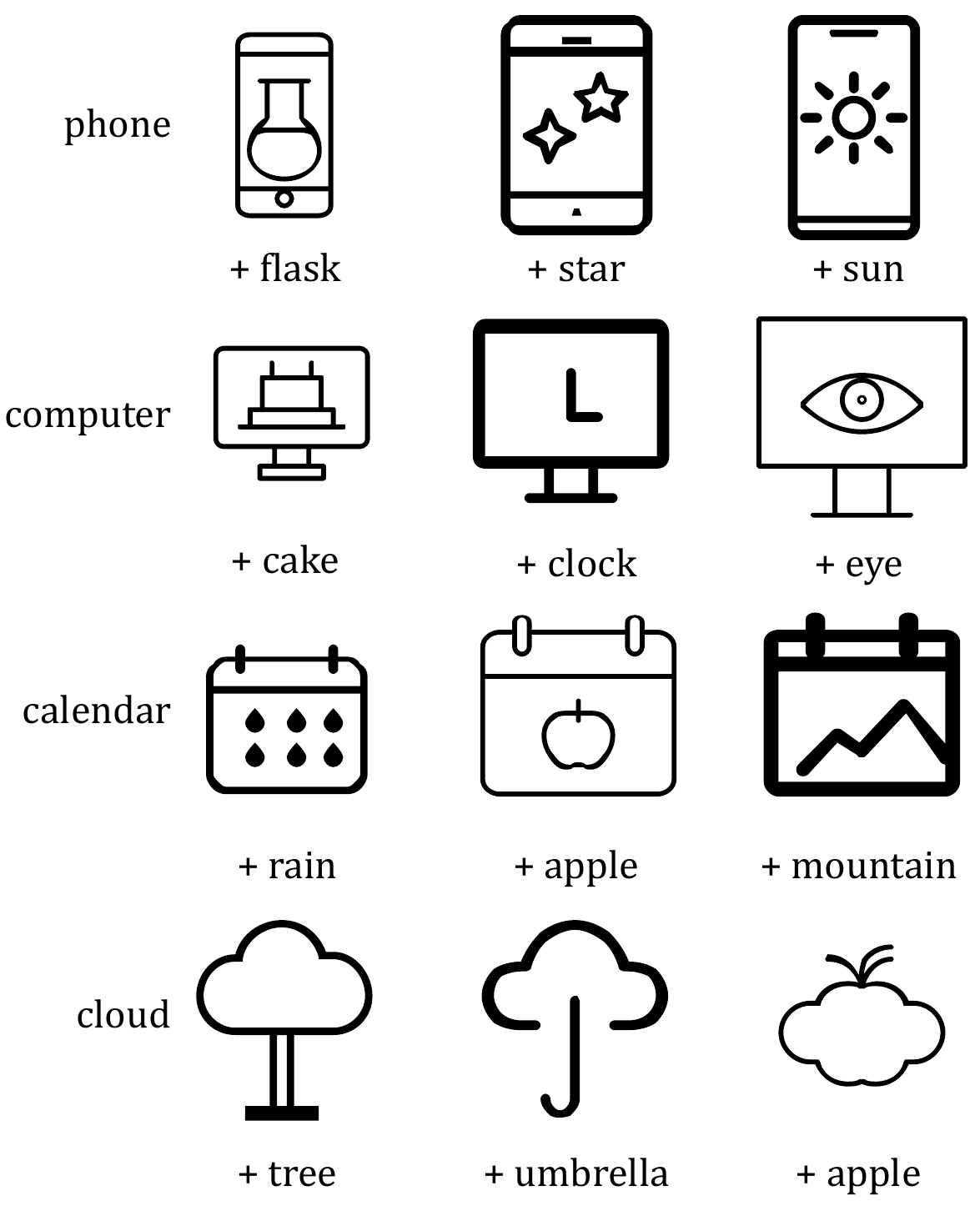}
    \vspace{-3mm}
    \caption{\sysName~learns to produce creative icons by combining semantics of different text prompts.}
    \vspace{-3mm}
\label{fig:text_combination}
\end{figure}

Text-to-image models demonstrate an impressive ability to combine textual inputs, allowing them to generate novel concepts that do not exist in the training data, such as ``\texttt{avocado chair}'' generated by DALL·E~\cite{ramesh2021zero}. In our experiments, we find that our model can also produce innovative and creative combinations, as shown in Figure~\ref{fig:text_combination}.

\subsection{Icon Design Auto-Suggestion}
\begin{figure}[t]
\centering
    \vspace{-2mm}
    \includegraphics[width=7cm]{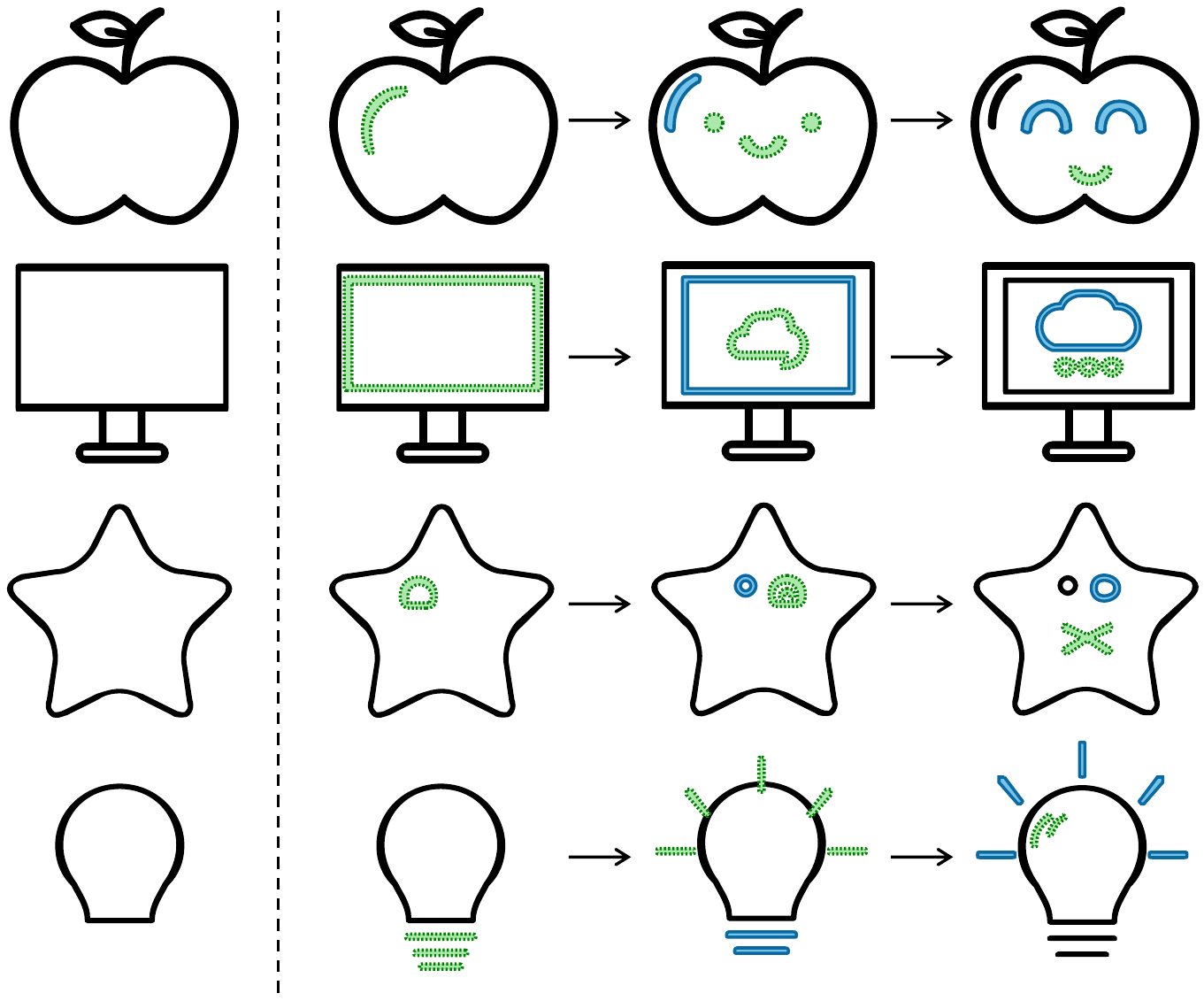}
    \vspace{-3mm}
    \caption{\sysName~is able to suggest subsequent paths for the users to design icons, with significantly boosted efficiency. We highlight paths suggested by \sysName~with the green color and paths drawn by the users with the blue color. Even if users diverge from the suggested route, \sysName~still predicts subsequent paths that are compatible with the uses' chosen paths.}
    \vspace{-3mm}
\label{fig:auto_suggestion}
\end{figure}

One advantage of automated icon generation is to support both designers and non-specialists in expressing their creative ideas.
A desired feature of such an automated system is the ability to suggest possible placements for subsequent paths on a canvas, which would significantly improve work efficiency and productivity. Relying on the autoregressive transformer, our trained \sysName~is able to predict the next path that users may choose in their icon creation processes (see Figure~\ref{fig:auto_suggestion}). Please refer to the project page for the video demonstrations of our auto-suggestion system for SVG icon design.

\section{Conclusion}
We have introduced \sysName, an autoregressive transformer-based method proficient at generating vector icons from textual descriptions. \sysName~stands out from both image-based methods that combine text-to-image generation and image vectorization, and language-based techniques that treat SVG scripts as natural languages.  Comprehensive experiments showcase the effectiveness and flexibility of \sysName~in terms of generation quality, diversity, text-icon alignment, and wide applicability. 
\begin{figure}[b]
\centering
    \vspace{-2mm}
    \includegraphics[width=6cm]{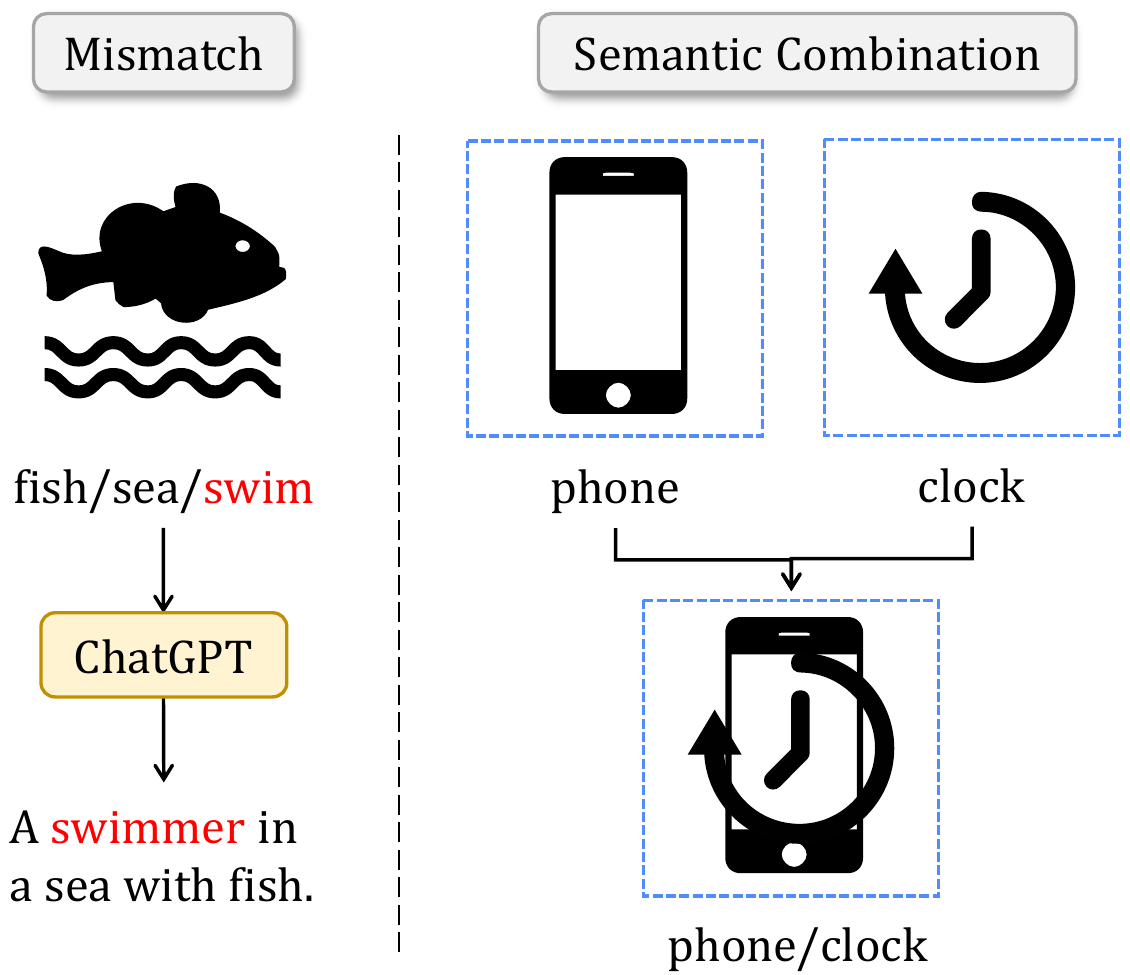}
    \vspace{-3mm}
    \caption{Limitations contain text-SVG mismatches (left panel) and suboptimal semantic icon combination (right panel).}
    \vspace{-3mm}
\label{fig:limitaion}
\end{figure}

\sysName~exhibits impressive SVG icon synthesis performance by exploiting the sequential nature of SVG scripts and unifying non-autoregressive and autoregressive modeling, but it is not without limitations (see Figure~\ref{fig:limitaion}). First, natural language phrases and sentences generated by ChatGPT may inadvertently result in text-SVG mismatches. This problem can be mitigated by using a higher-quality SVG dataset with accurate text annotations or by human filtering.
Second, the semantic icon combination performance by \sysName~may not be as remarkable as text-to-image generation, because most icons in the \textit{FIGR-8-SVG} dataset contain a single object situated at the center, occupying a significant portion of the space. We believe proper data augmentation such as scaling and merging SVG data to create new icons has the potential to notably improve the combination performance.
Finally, \sysName~is restricted to black-and-white icon generation, but it could potentially be expanded to yield multicolored icons or more general SVG content (e.g., clip art) with proper modifications. 

\bibliographystyle{ACM-Reference-Format}
\bibliography{ref}

\end{document}